\documentclass[3p]{elsarticle}
\usepackage{hyperref}
\usepackage[export]{adjustbox}
\usepackage{amsthm}
\usepackage{float}
\usepackage{soul}
\usepackage{color}
\journal{Journal of \LaTeX\ Templates}
\begin{document}
\begin{frontmatter}
\title{A Novel Fully Annotated Thermal Infrared Face Dataset: Recorded in Various Environment Conditions and Distances From The Camera}

\author[1]{Roshanak Ashrafi}\ead{rashrafi@uncc.edu}
\author[1,2]{ Mona Azarbayjani}
\author[3]{ Hamed Tabkhi}

\address{University of North Carolina at Charlotte}
\address[1]{Department of Civil and Environmental Engineering, The William States Lee College of Engineering} 
\address[2]{Department of Architecture, The College of Arts and Architecture}
\address[3]{Department of Electrical and Computer Engineering, The William States Lee College of Engineering}

\begin{abstract}
Facial thermography is one of the most popular research areas in infrared thermal imaging, with diverse applications in medical, surveillance, and environmental monitoring. However, in contrast to facial imagery in the visual spectrum, the lack of public datasets on facial thermal images is an obstacle to research improvement in this area. Thermal face imagery is still a relatively new research area to be evaluated and studied in different domains.The current thermal face datasets are limited in regards to the subjects' distance from the camera, the ambient temperature variation, and facial landmarks' localization. We address these gaps by presenting a new facial thermography dataset.\color{black}This article makes two main contributions to the body of knowledge. First, it presents a comprehensive review and comparison of current public datasets in facial thermography. Second, it introduces and studies a novel public dataset on facial thermography, which we call it Charlotte-ThermalFace. Charlotte-ThermalFace contains more than10000 infrared thermal images in varying thermal conditions, several distances from the camera, and different head positions. The data is fully annotated with the facial landmarks, ambient temperature, relative humidity, the air speed of the room, distance to the camera, and subject thermal sensation at the time of capturing each image. Our dataset is the first publicly available thermal dataset annotated with the thermal sensation of each subject in different thermal conditions and one of the few datasets in raw 16-bit format. Finally, we present a preliminary analysis of the dataset to show the applicability and importance of the thermal conditions in facial thermography. The full dataset, including annotations, are freely available for research purpose at https://github.com/TeCSAR-UNCC/UNCC-ThermalFace\color{black}
\end{abstract}
\end{frontmatter}

\section{Introduction}
The emergence of thermal imaging techniques has provided an excellent opportunity for contactless data gathering without interruption to occupants' activities. As elevated body temperature is an important indicator of a possible underlying physiological process, thermal imaging is a great non-intrusive tool for presenting that biological state. The non-invasive nature of thermal cameras has resulted in extensive utilization of these devices for detecting and diagnostic  purposes in several areas of medicine \cite{Lahiri2012MedicalReview,Antognoli2018AssessmentNeonates,Chakraborty2017HighCase}, and none-medical research such as fire safety\cite{Giitsidis2015,Szajewska2017,Ma2020}, transportation\cite{Chen2019,Miethig2019,Arabzadeh2019}, and building construction\cite{Dino2020,Gupta2018,Entrop2017}. 

Recent accelerated innovations in thermal imaging devices make it possible to utilize lower-priced thermal cameras to collect high-quality information that can improve several aspects of our lives. Facial thermography, in particular, is one of the most widely studied areas due to its proven practical applications. Facial thermal imagery has been successfully studied as an indicator of human identity\cite{Espinosa-Duro2013,Peng2016}, emotions\cite{Kopaczka2018,Goulart2019,Ordun2020}, and comfort \cite{Cosma2019UsingComfort,Lai2016AEnvironment,Lu2019ThermalBuildings} in several research. The current pandemic has also highlighted the facial thermography potential in the non-intrusive evaluation of human conditions more than before \cite{Jiang2020CombiningDevice,Al-Humairi2021}. 

While facial thermography has gained lots of recent attention, this is still a new research area that needs to be evaluated and studied in different domains. The lack of public datasets of facial thermal images is a significant obstacle to research improvement in this area. Currently, there are lots of RGB facial datasets available in diverse conditions \cite{Corneanu2016}, but the facial infrared image databases are limited and need improvement for several reasons. First, due to the lower quality of older thermal cameras, several current thermal datasets are of low resolution. They are not appropriate for use in sensitive areas, such as health-related applications \cite{Poster2021ADataset}. Second, most of the existing public datasets are not in the original raw format and are converted frames, which results in the loss of important information \cite{Kwasniewska2020Super-resolvedAnalysis}. The original thermal images are created utilizing 14-bit or 16-bit radiometric data that comprises information about the heat flux or temperature of each pixel. However, the majority of these datasets are transformed to lower bit resolutions, such as 8-bit RGB files, to make them smaller and more compatible with common imaging software. As a result of this process, temperature data is lost and cannot be retrieved later.\color{black}
Third, the current public datasets lack diversity in several aspects, such as face resolution and environmental properties. Most of the available datasets are appropriate for facial recognition purposes with no variation in environmental properties such as air temperature, relative humidity, and air speed.None of the current public datasets include data on controlled thermal variations, which is one of our top objectives in collecting this dataset. While some of the current datasets were recorded in uncontrolled thermal circumstances, resulting in temperature fluctuation, yet there is no information about the ambient temperature at the time of thermography recordings.\color{black}In addition, there are no publicly available datasets with varying distances from the camera\cite{Zhang2019SynthesisNetworks}, while several studies have indicated the importance of distance in the infrared thermography readings \cite{PlayaMontmany2021SpotThermography,Faye2016DistanceStudies,Vardasca2017TheTemperature}.In field thermography applications, when precision is required , the temperature variance induced by distance is substantial and needs to be considered. 
\color{black}
Fourth, while some recent research has demonstrated the great potential of using thermal cameras in smart buildings to predict occupant thermal sensations, none of them have made their datasets available for use in other projects. The improvements in thermal imaging technology enable the collection of data for use in the control systems of our smart buildings. Temperature data has been shown to be a viable variable for predicting occupant thermal preferences and controlling indoor environmental systems.While some of the current datasets were recorded in uncontrolled thermal circumstances, resulting in temperature fluctuation, there is no information about the ambient temperature at the time of thermography recordings.\color{black} However, there are currently no public datasets available that include the thermal sensation of the subjects in different thermal conditions. Finally, the majority of thermal datasets are either missing any landmark annotations or include just a few facial landmarks and the bounding box. Lack of manual facial landmark annotation makes the application of this datasets limited. Facial landmarks provide an extra supervisory signal and assist in the recognition of complex cases, in addition to making it possible to align the face during the face recognition process.\color{black}

\begin{figure}[h]
\centering
\includegraphics[width=\textwidth]{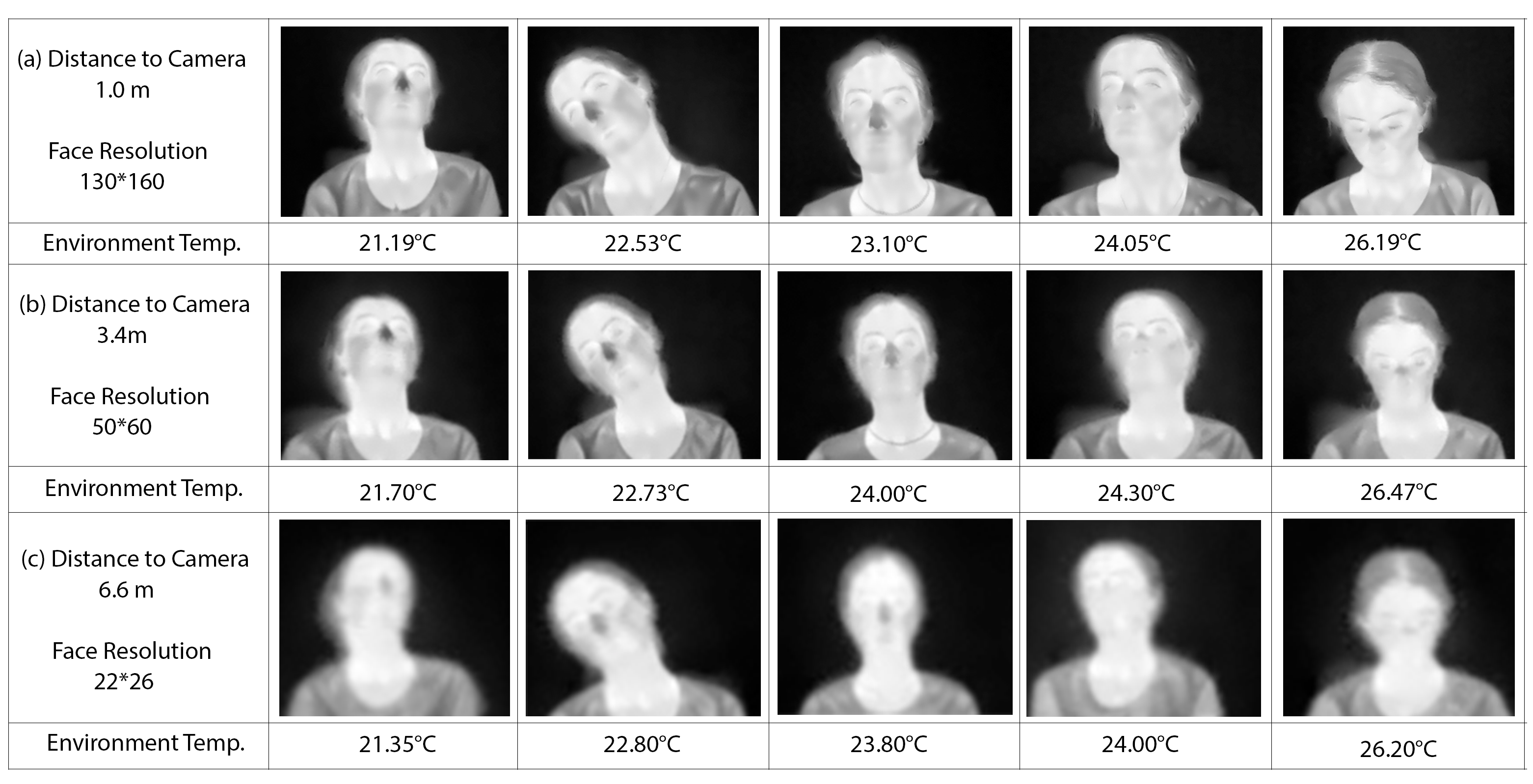}
\vspace{-15pt}
\caption{A Sample of Included Data Frames in Different Thermal Conditions}
\label{fig:sample}
\end{figure}

To address some of the limitations mentioned above, we present the Charlotte-ThermalFace dataset with unique properties. As Figure\ref{fig:sample}  shows through some sample images, our dataset includes thermal images of the same subjects in various thermal and physical conditions, including the environmental temperature, distance from the camera, and head position.  In contrast to the existing datasets that are recorded mostly at a fixed distance from the camera, our dataset includes a variation in distance, which has resulted in thermal images with several different resolutions. In addition, each subject is recorded in at least four different thermal conditions at all the specified distances.\color{black} We have included the environmental temperatures data in the dataset, which we present in this sample figure for each image frame. The first column of the figure shows the lowest temperature range images, which increase as we go to the left side columns. We can also observe in this figure that varying distances from the camera have resulted in a dataset with different thermal resolutions. We recorded the data frames at 10 relative distances from the camera, of which this sample figure includes the three main ones (closest, middle, and farthest).In addition to environmental data annotations, all the images are manually annotated with extensive facial landmark localization, which include 72 points for full or semi-profile positions, and 42 points for profile faces \cite{Bodini2019}.To avoid missing temperature data, our dataset is published in the original high accuracy recording format of Flir cameras, which is 16-bit raw data in the TLinear mode \cite{FLIRCompanyAIndustry}. This capability stores the thermal data for each pixel in the frames, and the users can read the temperature of each pixel independently.\color{black}
Based on the aforementioned information, the Charlotte-ThermalFace dataset is available publicly with the following main characteristics: 

\begin{itemize} 
\item We have captured approximately 10,000 infrared thermal images in varying thermal conditions, several distances from the camera, and changing head positions. We have also controlled the air temperature to change from 20.5°C (~69°F) to 26.5 °C(~80°F). Images are available in four different temperatures, 10 relative distances from the camera, starting at 1m (~3.3 ft) to 6.6m(~21.6 ft), and 25 head positions. 
\item The first public facial thermal dataset annotated with the environmental properties including air temperature, relative humidity, air speed, distance from the camera, and subjective thermal sensation of each person at the time. 
\item All the images are manually annotated with 72 facial landmarks. 
\item We are publishing the data in the original 16-bit radiometric TemperatureLinear format, which has the thermal value of each pixel. 
\end{itemize}

This study has two main contributions. First, it presents a comprehensive comparison of the current public datasets in facial thermography. Second, it introduces the Charlotte-ThermalFace public dataset on facial thermography with a brief investigation. The rest of this paper has three main sections. In section 2. literature Review, first, we look into the applications of facial thermography in different domains. Then we study and compare the existing publicly available thermal datasets with a brief description of each. In Section 3, we introduce our developed dataset and how we address some of the existing gaps by publishing our dataset. Section 4. provides a preliminary analysis of the dataset to show its applicability for future projects.

\section{Literature Review}
\subsubsection{Facial Thermography Applications}
A significant decrease in the prices of thermal cameras and their improved quality and resolution are paving the way for the utilization of these sensors in more research and real-world applications. The applications of biometric thermal images are defined in four categories: detection, monitoring, and recognition/identification \cite{Kristo2018AnMethods}. In this section, we have looked into the applications of facial thermography in these three major areas in detail. 

\textit{Face Detection}:

While variations in lighting conditions can easily degrade the performance of face detection algorithms, infrared thermography provides accurate results even in complete darkness. This quality has resulted in better performance of thermal images in comparison to visual images in pedestrian detection for autonomous vehicles. Their results show the Missing Rate (MR) when using Histogram of Oriented Gradients (HOG) features for visual images is \%73, which decreases to \%50 by using thermal image \cite{Chen2019c}. In regards to face detection, it is demonstrated that when employing visual photos for face identification, non-uniform illumination and fake faces may easily make cascade classifiers inoperable. However, the thermal frames do not have this downside. Through utilizing thermal images, machine learning-based face detection algorithms use classifiers such as AdaBoost and Support Vector Machine for considering facial and non-facial patches as positive and negative regions, which have been successful  \cite{Ma2017a}. Additionally, other research teams have conducted field experiments to evaluate and compare the proposed approaches to face detection using RGB photos and have demonstrated the benefits of utilizing thermal images to detect faces \cite{Ma2017}. Researchers have shown promising results using Haar-like features combined with a cascade of boosted tree classifiers \cite{Sumriddetchkajorn2009FaceLibrary}, and Faster R-CNN and YOLO \cite{Kowalski2021DetectionImages} as viable strategies for face detection in the thermal domain. It is important to mention that the accuracy of these face detection algorithms was shown to be significantly reduced when pictures are flipped and rotated.  \cite{Vukovic2019}. 

\color{black}
\textit{Face Monitoring}: 

The two major applications of facial monitoring is in medicine and building science, which are described below:

 \textit{Medical Facial Thermography}: The feasibility of using thermal scanners for reading skin temperature instead of traditional oral or rectal thermometers has been studied in several research \cite{Hughes2008Non-ContactEffectiveness}\cite{Selent2013MassSystems.}. The consensus, also supported by recent data, indicates that the temperature of the eye area has the highest temperature on the head, which makes it suitable for detecting fever \cite{Mercer2009}. It has also been shown that the side temple, inner eye area, and ear have shown the best correlation of internal body temperature \cite{Ng2005IsSARS}. At the same time, because these two areas are most affected by environmental temperature \cite{Silawan2018AScreening}, cheeks and nose temperature are promising measured facial areas for calibrating and offsetting the impacts of environmental factors. Thermal imaging has also been used for respiration rate detection by monitoring the temperature change frequency in the nasal area \cite{Alkali2013}. This approach can eliminate the adverse effect of the lighting condition; however, the relatively lower resolution of thermal cameras requires lower distances to the camera even in the recent research \cite{Cho2017RobustImaging}. The combination of RGB and thermal imaging has led to higher accuracy of respiratory rate monitoring \cite{Negishi2018StableScreening}\cite{Chen2019Rgb-ThermalMeasurement}\cite{Jiang2020CombiningDevice}. Due to the COVID-19 pandemic, the accuracy of infrared thermography in the medical sector has become more critical than ever. Some early studies have shown a few successful cases of using thermal cameras to detect febrile patients worldwide \cite{Lee2020WearingHospitals,Lee2020EffectiveHospital,Barnawi2021}. In this regard, the U.S. Food and Drug Administration (FDA) has enforced policies for Telethermograph systems during the COVID-19 pandemic based on a recent international standard (IEC 80601-2-59)\cite{Recommendations2020GeneralEpidemic}.

\textit{Environmental Monitoring}: Recent research has also proved thermal imaging to be a successful indicator of the building's occupants' thermal preferences. The infrared cameras can be installed at a distance from the occupant and capture the skin temperature by reading the pixel values of the desired facial regions, proving the feasibility of this technique with 94\% -95 \% accuracy when using FLIR A655 \cite{Ranjan2016ThermalSense:Imaging}. A real-time feedback system using FlirA35 thermal cameras was developed in 2018, analyzing both face temperature and occupants' position \cite{Metzmacher2018Real-timeAssessment}. A lower-cost could also replace the previously mentioned expensive cameras and smaller infrared camera, Flir Lepton, with an acceptable accuracy of 85\% for predicting the skin temperature \cite{Li2018Non-intrusiveThermography}. The accuracy of thermal comfort prediction was compared between different monitoring devices, including air temperature sensors, wearable skin temperature wristbands, and thermal infrared cameras. The findings highlight a slight improvement in the prediction accuracy by adding physiological sensors to the environmental sensors. However, it questions the efficiency of using physiological sensors for this slight accuracy increase (\%3-\%4) \cite{Aryal2019SkinAssessment}. In another recent study in this area, Li et al. had successful experience monitoring and recording the skin temperature of two occupants simultaneously with two thermal camera nodes, while each camera captured some parts of the faces \cite{Li2019RobustCameras}.

\textit{Facial recognition}:  

Facial recognition is one of the most studied areas of facial thermography \cite{Kristo2018AnMethods}. Face recognition in the thermal spectrum has gained more attention after showing successful results of Long Wave Infrared Imaging (LWIR), which was even superior to the visible spectrum face recognition results\cite{Selinger2006Appearance-basedStudy}. The research in this area has been ongoing ever since through utilizing several image processing methods, which have proved to be promising even in the outdo or environments, where illumination and face alignment vary substantially \cite{Mendez2009FacePatterns}. Multi-modal face recognition was also introduced in 2005 by utilizing both thermal and RGB images to develop a three-dimensional model of the face, and its thermal texture map \cite{Kakadiaris2005MultimodalInformation}. Research has performed experiments involving data fusion of multi-spectral imagery and was able to confirm the improvement in face recognition rate from 0.789 to 0.870 as a result of image fusion\cite{Chang2006AnRecognition}. \color{black} Furthermore, dual camera setups were designed to be used in dynamic illumination conditions to switch between the visible and infrared spectrum and select the most confident result in different lighting situations \cite{Serrano-Cuerda2014SelectionMeasures}.

Emotion and expression detection as two other application of face recognition, which have also shown promising results \cite{Robinson2012TowardTechniques}. 
for detecting conditions such as anxiety, fear, and alertness \cite{Pavlidis2002}. The main reason for effectiveness of thermal images for this application is variations in the skin temperature while expressing different emotions , which had made thermal data an essential source of additional information to improve in the evaluation of facial expressions and emotions \cite{Nguyen2014AAnalysis}. This quality has helped with creation and robot, capable of communicating with children and recognition of their emotions including disgust, fear, happiness, sadness, and surprise. These emotions could be recognized with an 85\% accuracy rate using low-cost hardware and low-cost approaches for visual and thermal image processing \cite{Goulart2019VisualInteraction}. Another research has shown the increased expression prediction accuracy from \%89.5 when using visual images to \%93.7 by using thermal frames. Multi-modal fusion through utilizing thermal, visual, and depth domains have also been proved to be successful in emotion recognition through utilizing more sophisticated late fusion approaches, such as fuzzy inference systems and Bayesian inference \cite{Corneanu2016}. Intoxication detection is another application of facial thermography in the facial recognition category. Alcohol consumption results in abnormally dilated blood vessels and increased blood pressure. In the facial area, this biological behavior shows itself by increasing the temperature around the nose, forehead, and eye area \cite{PradeepKumar2020}. Despite several challenges, recent research has shown an average percentage accuracy of \%99.63 in identifying drunk individuals through thermal images \cite{Sancen-Plaza2020}.

In conclusion, we have shown that thermal images can be of great importance in several domains.
Although both visual and thermal images are influenced by the change in lighting and thermal conditions, the intensity of the change varies extensively based on the cameras' dynamic range. As a consequence, when using an visual camera, the change in pixel values in the facial regions may be close to 100\% of the dynamic range of the sensor. However, the thermal domain changes are much less than the dynamic range that a typical thermal camera can capture, which will result in the higher consistency of thermal images, even in varying thermal conditions \cite{Ma2017}. 
\color{black}

\begin{figure}[H]
\centering
\includegraphics[width=\textwidth]{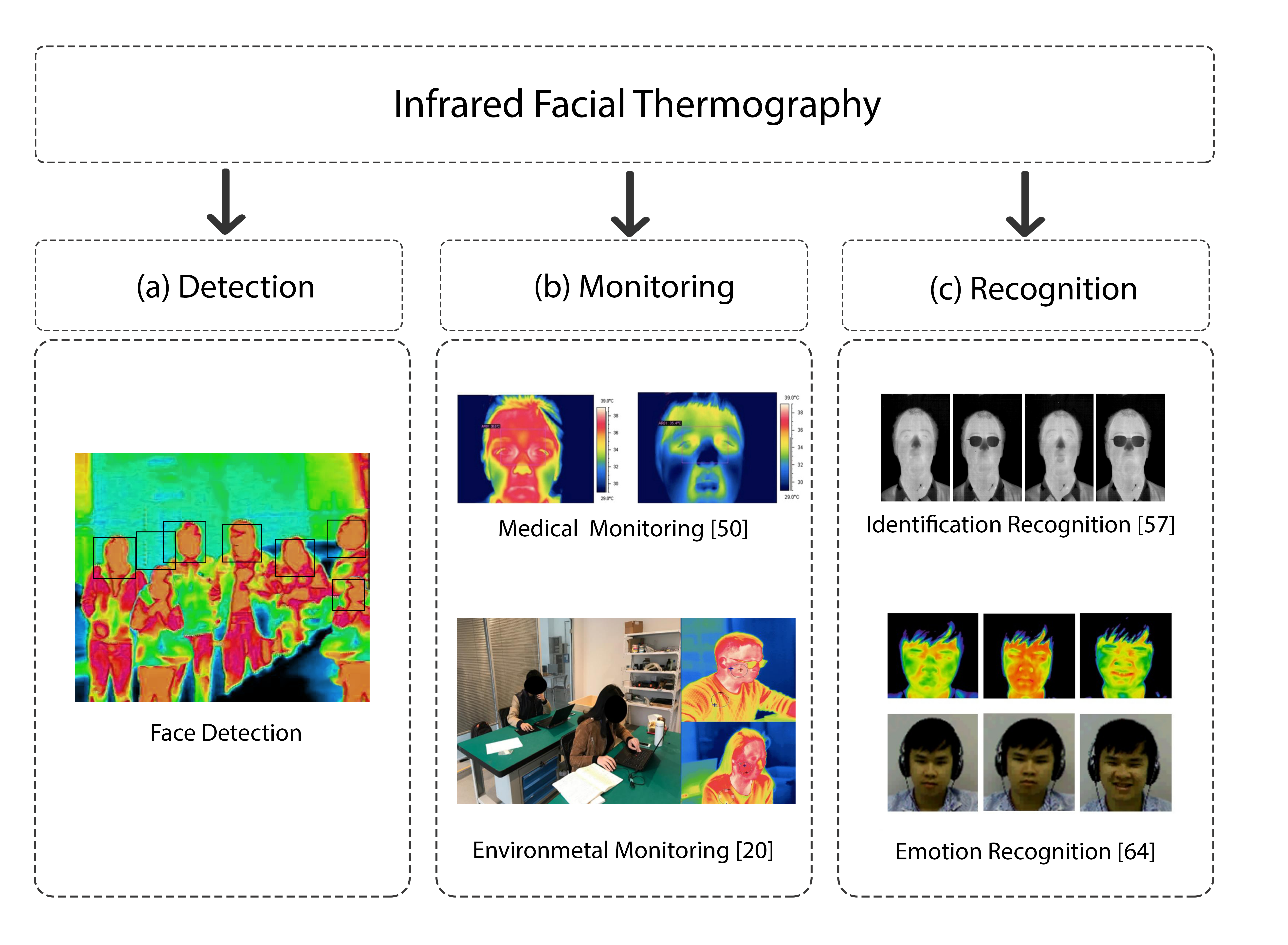}
\vspace{-15pt}
\caption{A Sample of Included Data Frames in Different Thermal Conditions}
\label{fig:sample}
\end{figure}

\subsection{Existing Facial Thermal Datasets}
Increased application of facial thermography in the mentioned sectors has highlighted the need for a comprehensive database for further studies. This section presents the current publicly available thermal face datasets and compares them to their key characteristics and limitations. This information is also available in Table\ref{table:lit} for better comparison. 

\begin{table}[h!]
\small
\begin{tabular}{ p{0.13\linewidth}c c c p{0.04\linewidth} p{0.0001\linewidth}p{0.04\linewidth} p{0.1\linewidth} p{0.06\linewidth} p{0.08\linewidth}p{0.08\linewidth}  } 
\hline
 Dataset & Thermal & RGB & Subjects &P&I&E&Resolution& Distance
 (cm) &Landmarks &Size(images)\\ 
\hline
 IRIS \cite{OTCBVS}&\(\surd\)  & \(\surd\)  &32& 11 & 5&3 &\( 320\times240\) & 183 &- & 4228 \\  
 IRIS-M3 \cite{Chang2006AnRecognition}&\(\surd\)  & \(\surd\)  &82& 25 & S*&1 &\( 640\times480\) & 120 &- & 2624 \\  
UND  \cite{Chen2003Visible-lightRecognition}&\(\surd\)  & \(\surd\)  &241&1 &3&3 &\( 320\times240\)& ? &-& 2624 \\ 
UH  \cite{Buddharaju2007}& \(\surd\)  &   &138&5&1&1-5 &\( 640\times512\)& ? &- & 7590 \\
{ FSU} \cite{Srivastava2001SpectralRecognition}&\(\surd\)  &  &10&0-20&1&0-20 &\( 320\times240\)& ? & & 234\\ 
{ Carl} \cite{Espinosa-Duro2013}&\(\surd\)  & \(\surd\) &41&1&3&1 &\( 160\times120\)&135&-& 7,380 \\ 
{ NVIE} \cite{Wang2010AInference}&\(\surd\)  & \(\surd\) &~215&1&3&7&\( 320\times240\)& 75 &-& 234\\ 
{KTFE}\cite{Nguyen2014AAnalysis}&\(\surd\)  & \(\surd\) &26&1&1&7&\( 320\times240\)& 85 &- & 126 GB\\ 
{ARL-MMFDV1}\cite{Hu2016AResearch}&\(\surd\)  & \(\surd\) &60&1&1&videoe&\( 640\times480\)& 250, 500, 750 &6& ?\\
{ARL-MMFDV2} \cite{Zhang2019SynthesisNetworks}&\(\surd\)& \(\surd\) &51&1&1&video&\( 640\times480\)& 250 &6& ?\\
{ULFMT}\cite{ShojaGhiass2018UniversiteUL-FMTV} &\(\surd\)  & \(\surd\) &236&1&1&7&\( 640\times512\)& 100 &-& ?\\
{Eurocom}\cite{Mallat2018AVariations}&\(\surd\)&\(\surd\)&50&4&5&7&\( 160\times120\)& 150 &-& 4200\\
{Tuft}\cite{Panetta2020ASystems}&\(\surd\)&\(\surd\) &113&9&2&5&\( 336\times256\)& 150 &-&10,000\\
{RWTH}\cite{Kopaczka2019ALabels}&\(\surd\)  & &90&9&2&8&\( 1024\times768\)& 90 &68& 10,000\\
{ARL-VTF}\cite{Poster2021ADataset}&\(\surd\)  &\(\surd\) &395&3&1&2&\( 640\times512\)& 210 &6& 500,000\\
{Sejong-A}\cite{Cheema2021}&\(\surd\)  & \(\surd\) &30&1&1&13&\( 768\times756\)& 200 &-& 1,500\\
{Sejong-B}\cite{Cheema2021}&\(\surd\)  & \(\surd\) &70&5-15&1&13&\( 768\times756\)& 200 &-& 23,000\\
{ I2BVSD}\cite{Dhamecha2013}&\(\surd\)  & \(\surd\) &75&1&1&7&\( 720\times576\)& ? &-& 681\\
{Sober-Drunk Database}\cite{Koukiou2012}&\(\surd\)  & &41&1&1&2&\( 128\times160\)& 30 &-& 4,100\\
{PUCV-DTF}\cite{Hermosilla2018}&\(\surd\)  &-&46&1&1&4&\( 640\times480\)& ? &-& 11,500\\
{TFW}\cite{Kuzdeuov2021}&\(\surd\)  &  \(\surd\) &147&S*&S*&video&\( 464\times348\)& S* &5& 9,982\\
{SpeakingFaces}\cite{Abdrakhmanova2021SpeakingFaces:Streams}&\(\surd\)  & \(\surd\) &142&9&1&S*&\( 464\times348\)& 100 &-& 4,581,595\\
{TIV}\cite{RolandMiezianko}&\(\surd\)  & &20&3&S*&3&\( 320\times240\)& ?&-& 21,676\\
 \hline
\end{tabular}
\label{table:lit}
\caption{ Public Available Datasets, P: Position, I: Illumination, E/D: Expression/Disguise,S*: Several}
\end{table}

 \textbf{IRIS}: The IRIS (Imaging, Robotics, and Intelligent Systems Lab) database contains 4228 pairs of images in both thermal and visible domains from 32 subjects, captured simultaneously. Similar to the Equinox database, each subject demonstrated three facial expressions: smiling, frowning, and surprised in five different illumination settings. This database also includes a 4-second video in each scenario at 10fps as the subjects pronounce the vowels. Each scenario is captured in 11 different positions by rotating the camera 36 degrees for each modality, at a fixed distance of 183 cm from the subject \cite{OTCBVS}.

\textbf{IRIS-M3}: This upgrade of the IRIS lab database contains 2624 pairs of images from 82 subjects in multi-band spectrum information, including one thermal and 25 visual bands per participant, in both outdoor and indoor environments. Another significant contribution of this upgrade is a diverse ethnic collection of Caucasian, Asian, Asian Indian, and African participants, with 24\% female and 76\% male participants. However, the database includes only one variation in facial expression and one frontal position toward the camera, while the distance from the camera is 1.2 meters for all the images \cite{Chang2006AnRecognition}.

\textbf{UND}: The University of Notre Dame database had published this database, which includes 2492 image pairs from 241 subjects, which is one of the most significant subject populations in the thermal/visual datasets. Each subject demonstrates two facial expressions of smiling and a natural face under three different indoor lighting conditions. Another unique feature of this dataset is capturing images in 4 sessions throughout the month. However, there are only four image pairs per subject \cite{Chen2003Visible-lightRecognition}. 

\textbf{UH}: The University of Houston database contains 7590 images of 138 subjects, captured in the Mid Wave MW thermal IR domain. There are 55 images of each subject in the database, demonstrating different facial expressions and arbitrary facial positions \cite{Buddharaju2007}.

\textbf{FSU}: The State University of Florida database contains 234 images from 9 subjects in thermal infrared in 7-14 \(\mu\)m spectral range information, which includes 25 visual bands and one thermal per participant. The images include varying angles and facial expressions for all the subjects. The dataset is at \( 320\times240\) resolution and in 8-bit BMP format \cite{Srivastava2001SpectralRecognition}.

 \textbf{Carl}: This dataset includes a total of 7,380 images that were recorded simultaneously with visible, near-infrared, and thermal sensors. 41 subjects participated in this study under three different illuminating conditions: natural, infrared, and artificial. The snapshots are all in frontal face position with neutral facial expressions. The Thermographic camera TESTO 880-3, equipped with an uncooled detector, captured both thermal and visible images. The images are captured in four separate sessions in which were two days apart \cite{Espinosa-Duro2013}. 

\textbf{NVIE}: The Natural Visible and Infrared Facial Expression database mainly focuses on capturing different facial expressions among several subjects. The dataset is from 215 subjects and includes two sub-databases: (1) Spontaneous and (2) Pose. The spontaneous dataset contains sequences from starting an expression to the final frame, and the Pose dataset contains only the last expression in both the visual and thermal spectrum. The researchers have captured all the images under three different indoor illumination settings \cite{Wang2010AInference} .

\textbf{KTFE}: Kotani Thermal Facial Emotion dataset includes simultaneous thermal and visible images in seven spontaneous emotions, including neutral, anger, happiness, sadness, fear, disgust, and surprise. 26 subjects participated in the study, with an age range of 11-years-old to 32-years-old. An infrared camera, the NEC R300, was used for capturing both thermal and visual videos. The dataset includes 126 gigabytes of visible and thermal facial emotion facial expression data frames. This database is one of the few datasets that mentions the recording room air temperature, kept between 24 °C and 26 °C \cite{Nguyen2014AAnalysis}. 

\textbf{ARL-MMFD}: Army Research Laboratory Multi-Modal Face Database was recorded simultaneously by a long-wave infrared camera and a visible spectrum camera. The utilized polymetric sensor is capable of recording geometric and textural facial details. The dataset included 60 subjects in the first published version\cite{Hu2016AResearch} that was then extended to 111 subjects \cite{Zhang2019SynthesisNetworks}. The researchers have used LWIR polarimetric for capturing images in both datasets, which are named Volume1 and Volume2. The expression change in the subjects' faces was created by counting out loud numerically. The first volume dataset is the only dataset collected at three different distances from the camera, which are 2 m, 5 m, and 7.5 m. However, the second volume is captured only at a single range of 2.5 m from the camera.

\textbf{ULFMT}: Université Laval Face Motion and Time-Lapse Video Database is recorded within four years from 238 subjects. This experiment includes different facial poses and expressions, ethnicities, ages, and sexes. The researchers have recorded the images in 4 different spectrum ranges with different cameras, including a Jenoptik camera for LWIR, a Phoenix Indigo IR camera produced by FLIR for MWIR, a CMOS made by Goodrich for SWIR, and a standard CCD made by Much for the NIR/Visible spectrum. The subjects had changed their facial expression arbitrarily and changed their head position from full-frontal face to complete profile for the video frames captured at 30 fps for 10 seconds \cite{ShojaGhiass2018UniversiteUL-FMTV}.

\textbf{Eurocom}: The dataset includes a total of 4200 images from 50 subjects, captured simultaneously in the visible and thermal spectrum. The dataset contains six illumination settings: ambient light, rim light, key light, fill light, all lights on, all lights off; seven expressions: neutral, happy, angry, sad, surprised, blinking, yawning; four head positions: up, down, right at 30°, left at 30°; and occlusion: eyeglasses, sunglasses, cap, mouth occluded by hand, eye occluded by hand. The thermostat temperature was at the average temperature of 25°C for the test room. The thermal camera is a Flir Duo with an uncooled VoX microbolometer and a thermal resolution of 160x120 pixels \cite{Mallat2018AVariations}. 

\textbf{Tufts}: The Tufts University database focuses on capturing images in various modalities from subjects in 15 different countries, genders, ages (4–70 years old), and ethnicities. A total of 10,000 images from 113 subjects had participated in the dataset. Images are under different scenarios of simultaneous visible and thermal images, near-infrared (NIR) images, a recorded video, a computerized facial sketch using the FACEs software, and 3D images of all the subjects. The researchers have utilized FLIR Vue Pro camera for recording the thermal images. The facial expressions for each subject are neutral, with a smile, eyes closed, an exaggerated shocked expression, and wearing sunglasses \cite{Panetta2020ASystems}. 

\textbf{RWTH}: The RWTH Aachen University database contains 2935 images in the thermal domain from 90 subjects. The subjects' distance from the camera is 90 centimeters. Each subject demonstrates six facial expressions of contempt, disgust, anger, fear, surprise, sadness, happiness, and a neutral face. The neutral scenario is captured in 9 different positions by rotating the head in vertical and horizontal positions. The database is a manually annotated dataset with 68 facial landmark points, emotions, and positions\cite{Kopaczka2019ALabels}.

 \textbf{ARL-VTF}: DEVCOM Army Research Laboratory Visible-Thermal Face Dataset (ARL-VTF) presents 500,000 images from 395 subjects, captured simultaneously with spectrum information. The dataset was recorded with a long wave infrared LWIR camera and three visible spectrum cameras. The FLIR Boson uncooled VOx microbolometer camera captured the thermal images with a spectral band of 7.5 µm to 13.5 µm. The database includes a variation in facial expressions created by counting out loud the numbers and facial positions created by rotating the head from left to right. The distance from the camera was 2.1 meters for all the images. The database is annotated with face bounding boxes and 6 points of facial landmarks, including left eye center, right eye center, the base of the nose, left mouth corner, right mouth corner, and center of the mouth  \cite{Poster2021ADataset}.

\textbf{Sejong}: Sejong is a recent multi-modal disguise face database, which contains images recorded in four modalities including visible, infrared, thermal, and visible-plus-infrared. The database contains two subsets and subset B  were captured one year after the first one (subset A). Subset-A has 30 participants (16 men and 14 females) and the total of 1,500 images, whereas Subset-B contains 70 subjects (44 males and 26 females) with the total number of 23,000 images. The highlight of this database is the add-on images that were captured in all four modalities. In the thermal images as they are not a part of the human body, disguise add-ons have a lower temperature than human skin and so seem darker than human skin or hair. In addition, subset B contains five to fifteen different poses for each subject. For the recordings the camera box were placed in the fixed distance of two meters from the subject and the room temperature was kept at 25 ± 5 ◦C.  The thermal images were captured by a Therm-App camera with the resolution of 768x756 pixels \cite{Cheema2021}. 

\textbf{I2BVSD}: The IIITD In and Beyond Visible Spectrum Disguise (I2BVSD) face database includes disguised/obfuscated face images in both thermal and visual spectrum. The database's disguise modifications are listed as follows. (1) No disguise: unambiguous appearance, (2) Hairstyles: wigs come in a variety of styles and hues. (3) Beard and mustache: many kinds of beards and mustaches (4) Eyeglasses: sunglasses and spectacles (5) Cap and hat  several types of caps, turbans, veils, and bandanas (6) Mask-related variation: disposable mask; and (7) Multiple variations: A mix of disguised accessories. The dataset includes 75 subjects, with one neutral and at least five disguised images for each subject. There are 681 images in each spectrum, which includes 6 to 10 images for each individual. The resolution of thermal images is 720 x 576 pixels.

\textbf{Sober-Drunk Database}: The dataset include images of drunk and sober individuals in thermal modality and was created at Electronics Laboratory, Physics Department, University of Patras, Greece. The images were captured with the FLIR Thermo Vision Micron A10 Model infrared camera with a resolution of 128x160 pixels. The dataset contains thermal images of both drunk and sober states for each individual. 41 individuals were included in this experimental method, 31 males and 10 females. The recording distance was 30 centimeters from the camera. The first 50 frames were acquired for each individual immediately before to initiating alcohol intake, and the second 50 frames were acquired 30 minutes after the fourth glass of wine was consumed. A total of 100 frames were captured for each participant, resulting in a database of 4100 pictures \cite{Koukiou2012}.

\textbf{PUCV-DTF}: The Pontificia Universidad Católica de Valparaíso-Drunk Thermal Face database is also a drunk classification dataset. 40 men and 6 women, with an average age of 24 years, have participated in this study. The images were captured with the FLIR TAU 2 thermal camera, which had a resolution of 640x480 pixels, a thermal sensitivity of 50 mK, and a wavelength range of 7.5–13.5 m. The dataset includes 250 images for each subject, which results in  11,500 total images. The room temperature and distance to the camera were not mentioned in the dataset publication \cite{Hermosilla2018}. 

\textbf{TFW}: Thermal Faces in the Wild (TFW) dataset contains thermal images of people both in outdoor and indoor environments with manually labeled bounding boxes and five-point facial landmarks (eye centers, nose tips, and mouth corners). The outdoor images were recorded in a variety of weather conditions, which included thermal photos acquired in both bright and cloudy conditions. The dataset has 9,982 frames and 16,509 labelled faces from 147 subjects. 5,112 images (5,112 faces) are recorded in a controlled setting and 4,870 image frames (11,397 faces) are collected outdoor. Occlusion, head position changes, various scales, face masks, diverse settings, and weather conditions are all part of the set. The FLIR T540 thermal camera, with a resolution of 464 x 348 pixels, a waveband of 7.5–14 m, a field of view of 24, and an iron color palette, was used to capture the datasets \cite{Kuzdeuov2021}. 

\textbf{SpeakingFaces}: SpeakingFaces dataset combines high-resolution thermal and visual spectral picture streams of fully-framed faces with audio recordings of each participant speaking about 100 sentences. The dataset includes 142 individuals, in nearly 13,000 instances of synced data from 9 different positions (3.8 TB). During a single trial, each individual participated in two different sessions. Subjects were requested to stay quiet and steady throughout the first session, while the operator captured visual and thermal video feeds from a sequence of nine collecting angles. The other session had the subject reading a sequence of words while visual, thermal, and audio data were recorded from the same nine camera angles. The images were recorded with  FLIR T540 thermal camera with a resolution of 464x348 pixels. Subjects were seated in a distance of one meter from the camera, and the room temperature was set at 25 degrees Celsius\cite{Abdrakhmanova2021SpeakingFaces:Streams}. 

\textbf{TIV}: Terravic Facial IR Database contains facial images in both thermal and visual spectrum. This database contains twenty individuals, each of whom has a unique set of frames with a variety of modifications, including front, left, and right orientations, indoor/outdoor setting and using  glasses or a hat. It contains 21676 thermal facial photos of 20 different individuals. The image frames were recorded with Raytheon L-3 Thermal-Eye 2000AS with a resolution of 320x240 pixels and delivered in the 8-bit gray scale JPEG format \cite{RolandMiezianko}. 

\color{black}
As presented, most of the current datasets include a diverse number of facial expressions, emotions, and head positions, which makes them suitable for facial recognition purposes. However, there is a substantial lack of facial landmark annotation in these datasets. Only RWTH and ARL-VTF have included manual facial annotations, while the ARL-VTF dataset contains only 6 main landmark points. Landmark detection is one of the most important tasks required for extracting biometric data from face thermal images. One main approach to detecting facial landmarks is utilizing visual images, which requires calibrating an RGB camera  and thermal camera together \cite{Wang2007}. Other approaches rely on the pixel value differences between the individuals' faces and their backgrounds. However, this approach does not work properly when the subject's head position changes \cite{Marzec2015}. As presented in RWTH dataset analysis \cite{Kopaczka2019ALabels} learning-based methods improve the facial landmark detection accuracy. However, there is currently only the RWTH dataset that includes a complete facial landmark annotation for 2935 images, which was inspired by similarly annotated datasets such as HELEN \cite{Le2012}. However, the data is gathered at a fixed distance of 90 centimeters from the camera and is published in 8 bit PNG format, which makes it loose some thermal information in the conversion process \cite{Kwasniewska2020Super-resolvedAnalysis}. The Charlotte-ThermalFace dataset provides all the images in a 16 bit TIFF format.
Furthermore, none of the currently available datasets covers controlled thermal variation, which is one of the main focuses of our data collection. Some of these datasets are recorded in uncontrolled thermal conditions, which results in temperature variation, but there is no information about the ambient temperature while thermography. Another value of our dataset is the ambient temperature and relative humidity variation, which is not been considered previously. We have recorded the thermal images for each subject in at least four different room air temperatures and included that information in the dataset annotations. 

In addition, the only dataset that has included distance in the facial thermography variation is the recent ARL-VTF dataset that has been developed by the DEVCOM Army Research Laboratory in 2021 and includes images at 3 different distances from the camera \cite{Poster2021ADataset}. However, this dataset does not include any controlled ambient temperature variation, and only 6 facial landmarks are tagged. 

Finally, as most of the presented datasets were recorded for computational purposes, and the human aspects such as the thermal sensation of the subjects were not considered. With the recent advances in thermography and its application in smart buildings \cite{Li2020HEATThermostat,Aryal2019ASensor,Lu2019ThermalBuildings} thermal imaging is going to be part of the future infrastructures. Our dataset would be a great help for researchers in this area. 

\color{black}
\section{Charlotte-ThermalFace: UNC Charlotte Thermal Face Database Overview}
As discussed in the previous section, the current facial thermography datasets are still very limited in quality and quantity and need improvement in many aspects. The facial thermal data is provided based on variation in environmental temperature, distance, and head position. We have gathered ~10000 infrared thermal images from 10 healthy subjects in varying thermal conditions, several distances from the camera, and different head positions. We have annotated the data with the ambient temperature, relative humidity, and air speed of the room at the exact time of capturing each image. In addition, 72 facial points are manually marked and added to the annotations.

UNCC-ThermalFace is the first publicly available thermal database annotated with the thermal sensation of each subject in different thermal conditions. It also enables radiometric enabled raw data frames. The radiometric option defines retaining the electromagnetic radiation in the data frame files\cite{FLIRCompanyAIndustry}. By enabling this option, our dataset provides information about the captured radiance in each pixel of the recorded images. All the data is in the original 16-bit radiometric raw format, with a thermal value for each pixel. Flir A700 has recorded the frames, which is one of the most recent Flir Systems' cameras \cite{FLIRSystemsFLIRSystems}. The thermal sensor is an uncooled microbolometer with the temperature range of 0°C to 650°C and accuracy of ±2°C (±3.6°F) or ±2\% of reading for ambient temperature 15°C to 35°C. The images are captured at 10 relative distances to the camera for each temperature range, as is shown in Figure \ref{fig:pos}. The original resolution of the thermal sensor is 640x480 pixels, and the resolution of the cropped facial area varies for each distance range.

\begin{table}[h]
\centering
\small
\begin{tabular}{  c c c c c c c c } 
\hline
 ID & Sex  & Age &Height(cm)&Weight(kg)&BMI(kg/m²)\\ 
 \hline
 1& Female  &34&168 & 69&24.4  \\ 
2& Male  &42&170 & 87&30.1 \\ 
3& Female  &30&170 & 54&18.7  \\ 
4 & Male  &33&173 & 70&23.4\\ 
5 & Female  &34&163 & 57&21.5 \\ 
6& Female  &35&171 & 89&29.4  \\ 
7 & Male  &32&168 & 70&24.8\\ 
8& Male  &34&183 & 78&23.3  \\ 
9& Female  &33&168 & 72&25.5  \\
10& Male  &27&175 & 72&23.5  \\
AVG& -  &33.4&170.9 & 71.8&23.3  \\
 \hline
\end{tabular}
\caption{Subjects' Information}
\label{table: subject}
\end{table}

\subsection{Data Collection Methodology}
The Office of Research Protections and Integrity UNCC-IRB183845 has approved this study. The data collection process took place in Jun 2021. We have collected the data in a one-day long session or two shorter sessions over two days based on the subject's preference. Participants are five males and five females, healthy adults. We made certain that the subjects did not have any thermoregulatory illnesses, heat intolerance, colds, flu, or infections. The participants did not use any makeup or facial cream and removed their glasses for the recordings. All participants wore light-colored short-sleeve shirts and pants. Two of the male participants had facial hair. Table \ref{table: subject} shows the information for each subject in detail. Each recording session is designed for capturing thermal recordings from one subject at ten relative distances from the camera and 25 different head positions in each interval. In addition, the temperature of each session is different from the previous session, which have resulted in several variations in the thermal condition. All the ten subjects have participated in at least four recording sessions.
\color{black}

After taking the informed consent, the researchers recorded the participant's age, gender, height, and weight. Adaptation is a common part of human-related thermal condition research, which helps subjects' bodies reach a steady metabolic rate and thermal state, as it is influenced by their prior activities and environments. Earlier research has suggested that the mean skin temperature and thermal sensation stabilize after 40 minutes of being exposed to a new thermal environment with a temperature difference of less than 10°C \cite{Jazizadeh2013ASensing,Jazizadeh2014User-ledBuildings,Balaji2013}. The adaptation time period is sometimes selected to be less than this period in similar previous research, such as 20 minutes\cite{Cosma2018ThermalCamera} or 30 minutes \cite{Choi2012InvestigationSensations}. As one purpose of our experiment has been the development of a facial thermography dataset, we have decided to choose a longer acclimatization period so we can be sure of the reliability of our dataset. The participants had entered the test room and stayed in a seated position for 60 minutes before the test, so their metabolic rate had reached a stable state, and any influence from the prior outdoor temperature was eliminated.\color{black} The experiment for each participant was a combination of five sub-sessions in an approximately one-hour time frame for each sub-session. The recording was initiated at a temperature of 21°C while the thermostat held the ambient temperature constant with a 1°C fluctuation allowance in each steady state session. Each subsequent session started at a 1.5°C higher temperature in the same one-hour time frame.

\begin{figure}[H]
\centering
\includegraphics[width=0.4\textwidth]{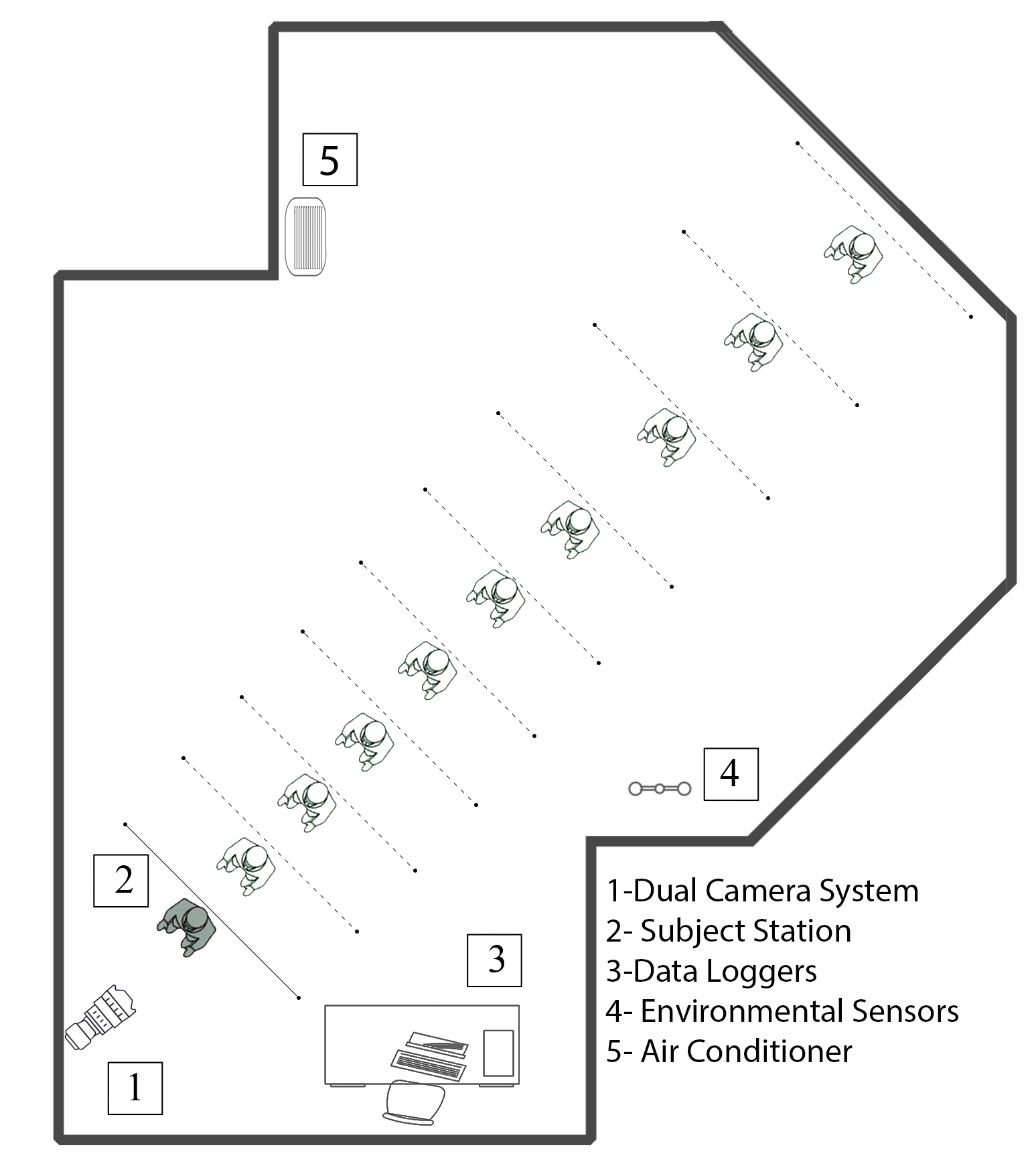}
\caption{Test Room and Recording Stations' Layout}
\label{fig:pos}
\end{figure}

Figure  \ref{fig:pos} displays the test room layout and the positioning of the following: (1)thermal camera, (2)subject's stations, (3)data loggers, (4)temperature and relative humidity sensors, and (5)air diffuser and airflow sensor. This figure shows that the temperature and humidity sensors are placed at the number(4) and mounted on a pole at three different heights (0.1, 1.1, and 1.7 meters). The recordings are at ten different relational distances from 1 to 6.6 meters, which this figure shows by number(2). Participants were seated in front of the camera during the study and changed their head positions as instructed while recording RGB and thermal frames pairs.

\begin{figure}[H]
\centering
\includegraphics[width=\textwidth]{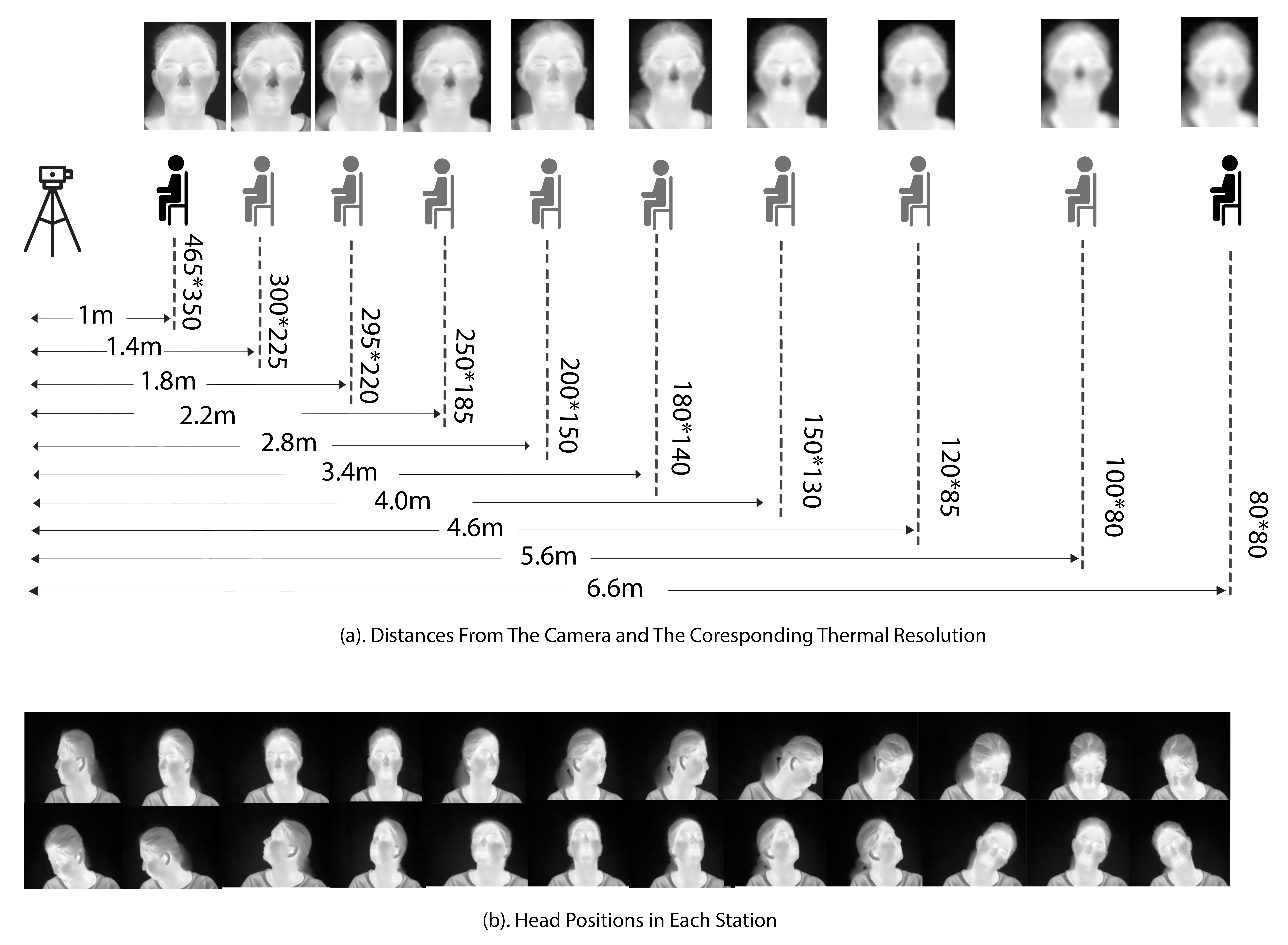}
\vspace{-20pt}
\caption{Distances From The Camera and Head Positions}
\label{fig:Head-Pos}
\end{figure}

The researchers performed the recordings in each station from the camera and in 25 different head positions instructed to the subjects (Figure \ref{fig:Head-Pos}). The user logged the thermal sensation in each station from the camera through a "Google Form" with three standard levels: cool, slightly cool, neutral, slightly warm, and hot. An approximate number of 1,000 frames were captured for each subject, with a total of 10,000 thermal frames for the dataset. Table \ref{table:vars} shows the recorded variables and recording sensors' information in more detail. The air temperature and relative humidity are measured with HOBO Pro v2 temperature/relative humidity data logger sensors, which were calibrated with ice water before the experiments. The airspeed is recorded at the air diffuser proximity with an average distance of 3 meters from the subject's station. The dataset is annotated with the face landmarks coordinates, distance to the camera, room air temperature, relative humidity, airflow, and subject thermal sensation. 

\begin{table}[h]
\centering
\small
\begin{tabular}{  c c c c c   } 
\hline
 Variable& Device Brand  & Model &Accuracy&Resolution    \\ 
 \hline
 Indoor Air Temperature& 	Onset  & S-THB-M008&+/- 0.21°C from 0° to 50°C & 0.02°C at 25°C \\ 
Relative Humidity&	Onset  & S-THB-M008& +/- 2.5\% from 10\% to 90\% RH&  0.1\% RH  \\  
Air Velocity& Fluke &922 Airflow Meter&±2.5\% of reading 10.00 m/s)& 0.001 m/s\\  
 \hline
\end{tabular}
\caption{List of Data Acquisition Devices}
\label{table:vars}
\end{table}

\subsection{Dataset Analysis}
This section presents a preliminary study of the gathered dataset. The goal of this evaluation is to investigate the applicability of the data. The analysis of the subjects' thermal sensation based on their skin temperature is not in the scope of this study and will be covered in our next publication as our objective in this section is to show the key statistical properties of the whole dataset. First, we analyze the main independent variables, including environmental temperature, relative humidity, and distance from the camera. Then, we look into the correlation of facial skin temperature in different facial areas with the environmental temperature and each other. The authors will also compare sample frames together to study skin temperature differences in diverse thermal conditions and among different subjects. The provided results for facial area correlation are based on our developed method for detecting facial landmarks. The complete explanation of this method is outside the scope of this paper and will be explained in detail in another relevant publication. In addition, we have not included the utilized visual images in the published dataset. Some recent research has succeeded in detecting facial thermal images without using RGB data \cite{Kowalski2021DetectionImages}. 

\begin{figure}[h]
\centering
\includegraphics[width=0.6\textwidth]{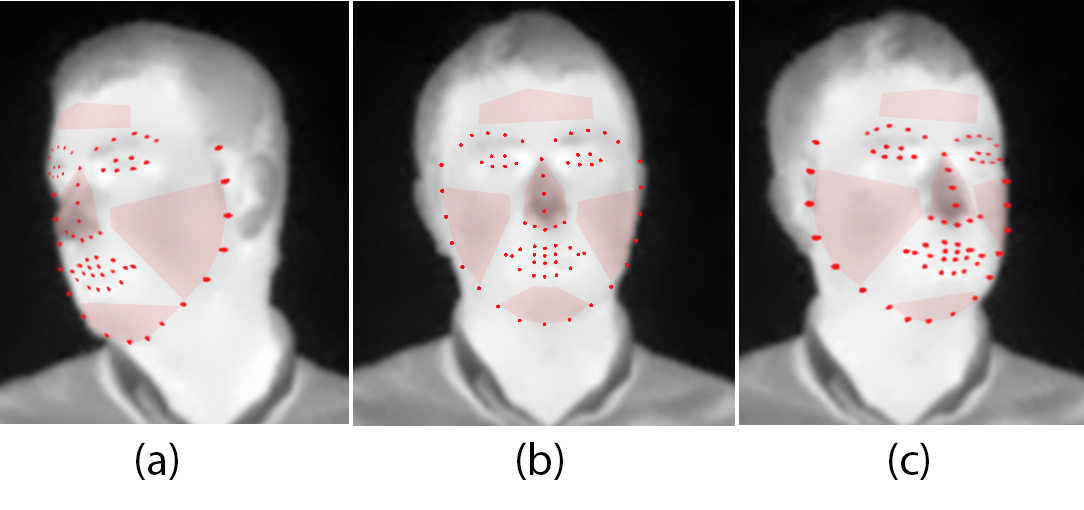}
\vspace{-10pt}
\caption {Identified Facial Areas Based on Landmarks In Three Different Facial Positions}
\label{fig:landmark}
\end{figure}

Since our dataset includes images with various resolutions, it would be a great fit to evaluate the prediction accuracy of such algorithms, which would be one of our future objectives. Here we provide a brief explanation of data preparation and the results’ analysis. We have first cropped the images to the facial area with some margins.  We have utilized the Dlib-based model for face recognition, which is based on a 29 convolutional layer in Residual Networks ResNet. This model is a version of the ResNet-34 network that works by removing some layers and reducing the number of filters per layer to half \cite{He2016DeepRecognition}\cite{DavisKing2017DlibLearning}. For locating the facial areas, the position of the facial landmarks has been transferred from the RGB image to the thermal image by calculating the homography matrix between the two frames, as is presented in detail by Negishi et al. \cite{Negishi2019InfectionAlgorithm}. As the covered distance is 1-6.6 meters from the camera and the facial area resolution gets as low as 25*30 pixels, the HRNet algorithm has been used to identify facial landmarks in the RGB images \cite{Wang2020DeepRecognition}. Four facial areas (nose, cheeks, forehead, and chin) have been selected based on the literature to be studied individually in detail \cite{Aryal2019SkinAssessment,Li2019RobustCameras}. \color{black}

The authors have utilized the estimated facial landmarks to define the selected facial areas, as shown in Figure \ref{fig:landmark}. Since the recorded frames are raw radiometric data, the value of each pixel presents the temperature of the pixel's representing area in the actual world. According to the Flir camera's datasheet, when using the TemperatureLinear (TLinear) option, the temperature of each pixel in degrees of Kelvin can be defined with Formula \ref{eq1}. This is also very important to set the emissivity of the measuring target to the correct number before the captures, which is 0.98 for the human skin. We have set the TLinear resolution is set to 100 mK(millikelvins) in our recordings \cite{FLIRCompanyAIndustry}.

\begin{equation} \label{eq1}
Skin Temperature = (Pixel Value/ Tlinear resolution  – 273.15K) 
\end{equation}

\section{Dataset Evaluation}
\begin{figure}[h]
\centering
\includegraphics[width=\textwidth]{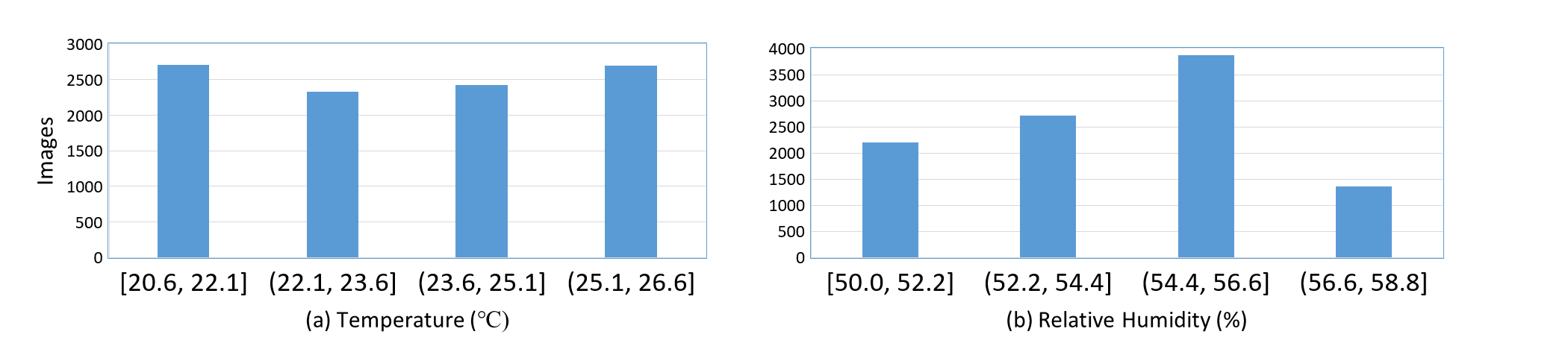}
\vspace{-15pt}
\caption {Covered Environmental Temperature and Relative Humidity in the Dataset}
\label{cover}
\end{figure}

First, the dataset frames' environmental properties are studied to ensure the proper coverage of different thermal conditions in the dataset. Figure \ref{cover} displays the (a)temperature and (b)relative humidity coverage based on the number of images for each range. The temperature range is between 20.6 °C and 26.6 °C, divided into four groups with a 1.5 °C range for each. This figure shows an approximately uniform distribution in the room (a)air temperature, which means the number of data frames in different temperature ranges is approximately the same. Although we have not controlled the (b)relative humidity in the recording process, controlling the air temperature has resulted in changes in relative humidity, which were recorded and added to the annotations. Additional statistical information about the environmental properties, including air temperature, globe temperature, relative humidity, and airflow, is provided in Table \ref{tab:env.}. Most of the sessions had zero airflows to ensure image consistency; however, the air conditioner was at 0.5 m/s airspeed at the diffuser proximity in extremely cold conditions.

\begin{table}[h]
\centering
\small
\begin{tabular}{  c c c c c c c c } 
\hline
  &Mean &Median&Range&Minimum&Maximum&Standard Deviation\\ 
 \hline
Air Temperature(°C)& 23.62&323.65&6.00&20.61&26.59&1.76 \\
Globe Temperature(°C)& 23.89&24.05&7.38&20.63&28.01&1.84\\ 
Relative Humidity(\%)&54.16&54.45&8.63&49.96&58.59&2.07\\ 
Air Speed(m/s) & 0.08&0.0&0.508&0.0&0.508&0.18\\ 
 \hline
\end{tabular}
\caption{Environmental Properties Measurement Results Overview}
\label{tab:env.}
\end{table}

\begin{figure}[H]
\centering
\vspace{-20pt}
\includegraphics[width=0.6\linewidth]{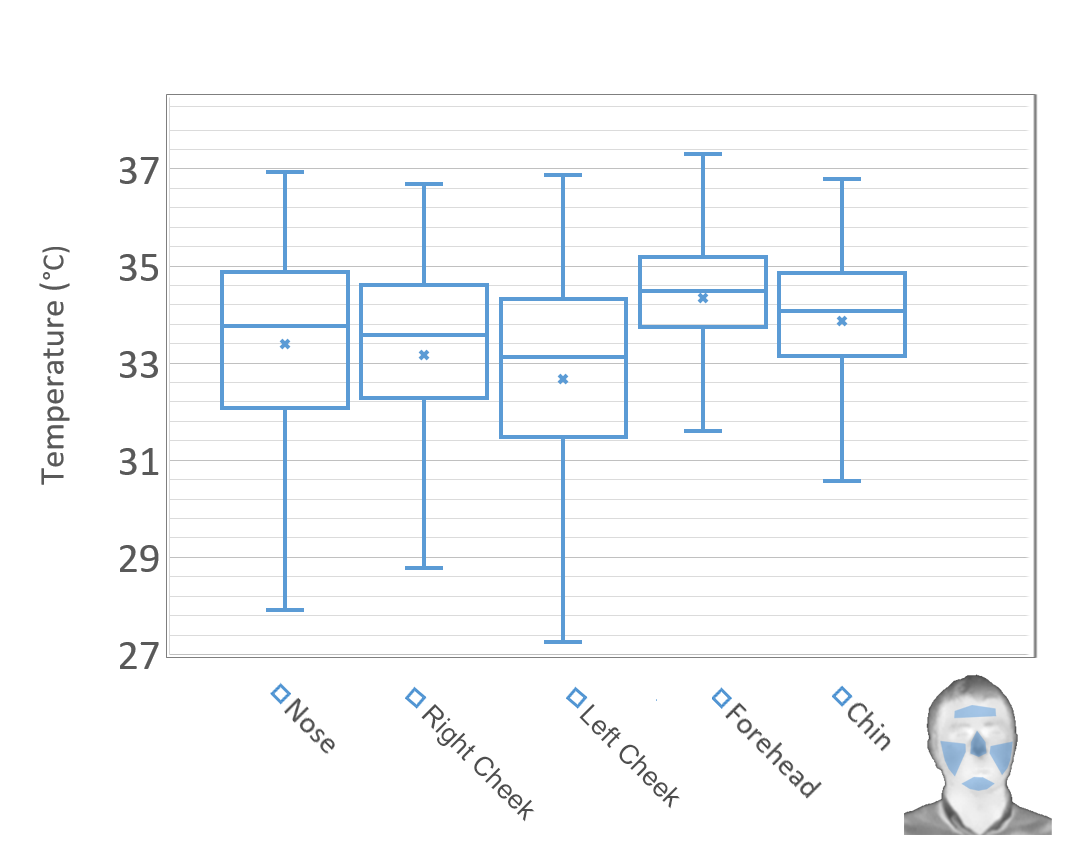}
\caption{Skin Temperature Range For Each Facial Location }
\label{fig:2}
\end{figure}

Furthermore, we have studied the changes in the skin temperature in each region for the whole dataset. The temperature of each pixel in the desired Region of Interest (ROI) is calculated by the equation\ref{eq1}, used to identify the average temperature in each facial area. Table  \ref{table:5} shows an overview of the dataset's skin temperature measurement results for all the recorded frames. As the table shows, the temperature in the forehead area has the highest mean value (34.3 °C), while other facial areas' mean values are closer together in the range of 33.17 °C to 33.82 °C. The low standard error numbers (0.02-0.03) in all facial regions demonstrate that a sample data set can be an acceptable representation of the entire dataset. In addition, the close amounts of mean and median numbers show the relatively symmetrical distribution of the data. Additionally, Figure \ref{fig:2} presents box plots for the temperature variation of each facial region based on the environmental temperature. As expected by the literature, the nasal area has the highest variation, and the forehead temperature shows the lowest variation.

\begin{table}[h]
\centering
\begin{tabular}{  c c c c c c c c c} 
\hline
 &Right Cheek &Left Cheek&Cheeks Average &Nose&Forehead&Chin\\ 
 \hline
 Mean&32.72&33.17&33.71&33.82&34.30&33.18\\
 Standard Deviation &2.18&2.01& 1.28&1.41&1.31&1.897\\ 
Standard Error&0.03&0.02&0.02&0.02&0.02&0.03\\ 
Median&33.15&33.57&33.86&34.07&34.47&33.55  \\ 
Sample Variance&4.76&4.03&1.63&1.98&1.71&3.56\\
 \hline
\end{tabular}
\caption{Skin Temperature Measurement Results Overview For All Subjects}
\label{table:5}
\end{table}

\begin{figure}[H]
\centering
\includegraphics[width=0.85\textwidth]{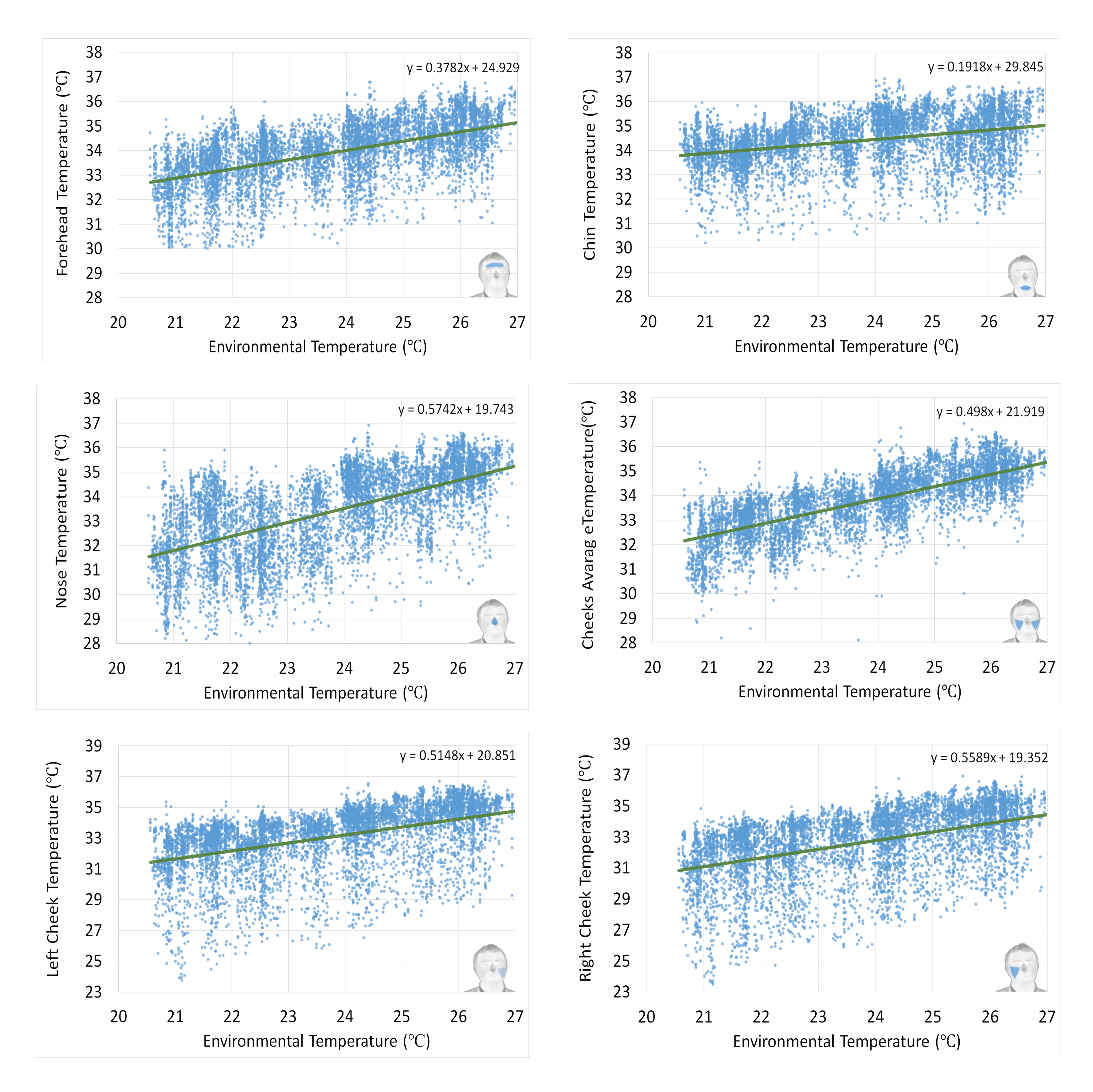}
\vspace{-10pt}
\caption{Correlation of Facial Locations With Room Temperature for All Subjects}
\label{fig:scatter}
\end{figure}

To have another holistic view of the changing patterns of different facial parts with room temperature, figure \ref{fig:scatter} shows the relationship between skin temperature and room temperature. The changing pattern of these charts shows that the temperature in all facial areas for the whole dataset has a positive weak linear association with the room ambient temperature. The  The \(R^2\) values for these linear patterns are chin 0.13, forehead 0.4, nose 0.4, cheeks' average 0.6, left cheek 0.25, and right cheek 0.25, which backs up this linear relation. The linear correlation values with the Pearson method are also presented in Table \ref{table:corr}, which will be further discussed.\color{black}As it is shown the forehead area has the lowest linear slope (0.1918) and the temperature range as the air temperature increases to the highest amount. On the other hand, the nose temperature shows the most significant change as the air temperature increases with a linear slope of 0.5742. These findings are in line with the literature \cite{Silawan2018AScreening} , highlighting that the nose area has the highest correlation with the room temperature. In contrast, the forehead area changes the least and can be a good indicator of internal body temperature. As shown in this figure, we have looked into both rights and left cheek temperature and their average temperature. The average cheek temperature is the average temperature of the left and right sides when both sides are visible in the thermal frame. If the head position is a full profile, the invisible cheek is excluded from the calculation. Therefore, although the number of data points for the right and left cheek is less than the other facial parts, the cheek average has the same data point as the other face areas. 

Furthermore, Table \ref{table:corr} presents the "Pearson Correlation Values" for different facial regions with the environmental temperature and the subjects' thermal sensations across the whole dataset. The Pearson correlation coefficient indicates the strength and direction of a linear relationship between different variables, and it would be a relevant factor for our investigation. Table \ref{table:corr} shows that the cheek average and nose area have the highest correlation with the environmental temperature, which is 0.76 and 0.67, respectively. On the other hand, the forehead area has the lowest correlation with the environmental temperature, which proves the applicability of this facial area as an indicator of core body temperature, as environmental properties less influence this area. This result is in line with the previous findings from the literature\cite{Silawan2018AScreening} on the correlation of different facial regions with the environmental temperature. In addition, we can see that the cheeks and the nose area are the best indicators of subjective thermal sensation, while the forehead area is not that suitable for that purpose.

\begin{table}[H]
\centering
\begin{tabular}{  c c c c c c c  } 
\hline
  &Right Cheek &Left Cheek&Cheeks Average &Nose&Forehead&Chin\\ 
 \hline
 Correlation with\\ Room Temperature&0.5 &0.5&0.76&0.67&0.35&0.55\\
 Correlation with\\      
 Thermal Sensation&0.42 &0.43&0.63&0.55&0.22&0.45\\
 \hline
\end{tabular}
\caption{Correlation of Facial Regions With Room Temperature and Thermal Sensation}
\label{table:corr}
\end{table}

\begin{figure}[H]
\centering
\includegraphics[width=0.8\textwidth]{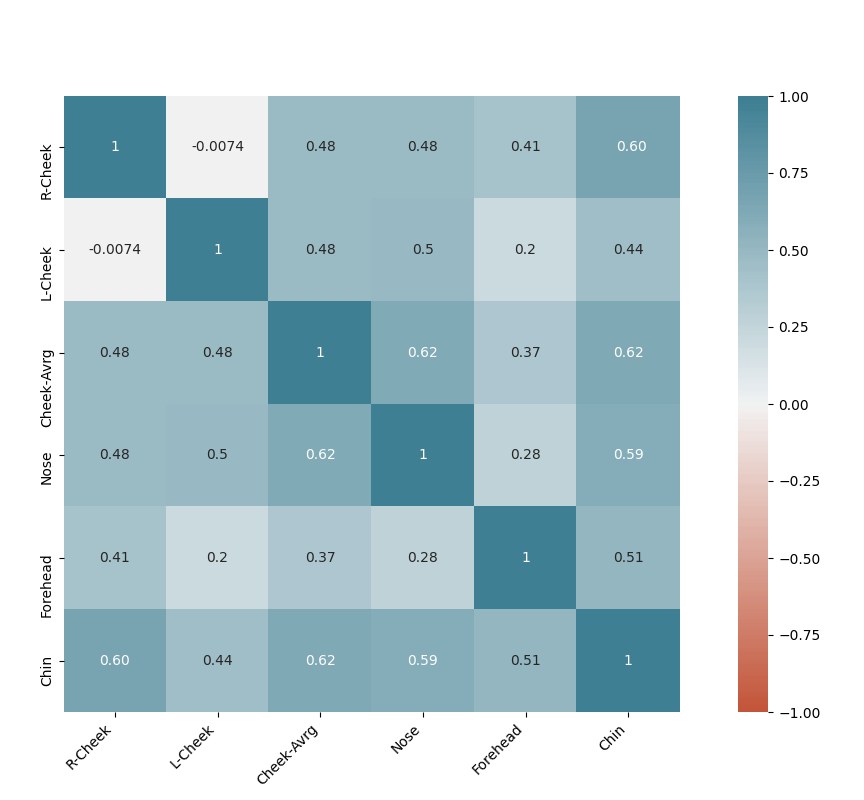}

\caption{Correlation of Different Facial Areas Among All Subjects}
\label{fig:correlation}
\end{figure}

The relationship between different facial regions among all the subjects is also calculated based on the Pearson Correlation Coefficient and visualized in a heat map figure. As Figure \ref{fig:correlation}  shows, most of the facial regions have a moderate to relatively strong correlation . The right and left cheeks show no correlation, which results from the change in the head positions that makes the two sides have different angles with the camera in most of the data frames. When the head is in a semi-profile position, one of the cheeks is right in front of the thermal camera, while the other side has an angel with the thermal camera, which makes the reading of that side lower than the actual temperature. As a result, the two facial sides can only have an approximately similar reading condition in the full frontal face position. This correlation heatmap also shows that the cheek area has the highest correlation with other facial regions, especially the nose and chin region (0.62). On the other hand, the forehead has the lowest correlation values with other facial regions, which was expected based on both literature and the correlation values of the forehead area with the environmental temperature from Table \ref{table:corr}.

\begin{figure}[h]
\centering
\includegraphics[width=\linewidth]{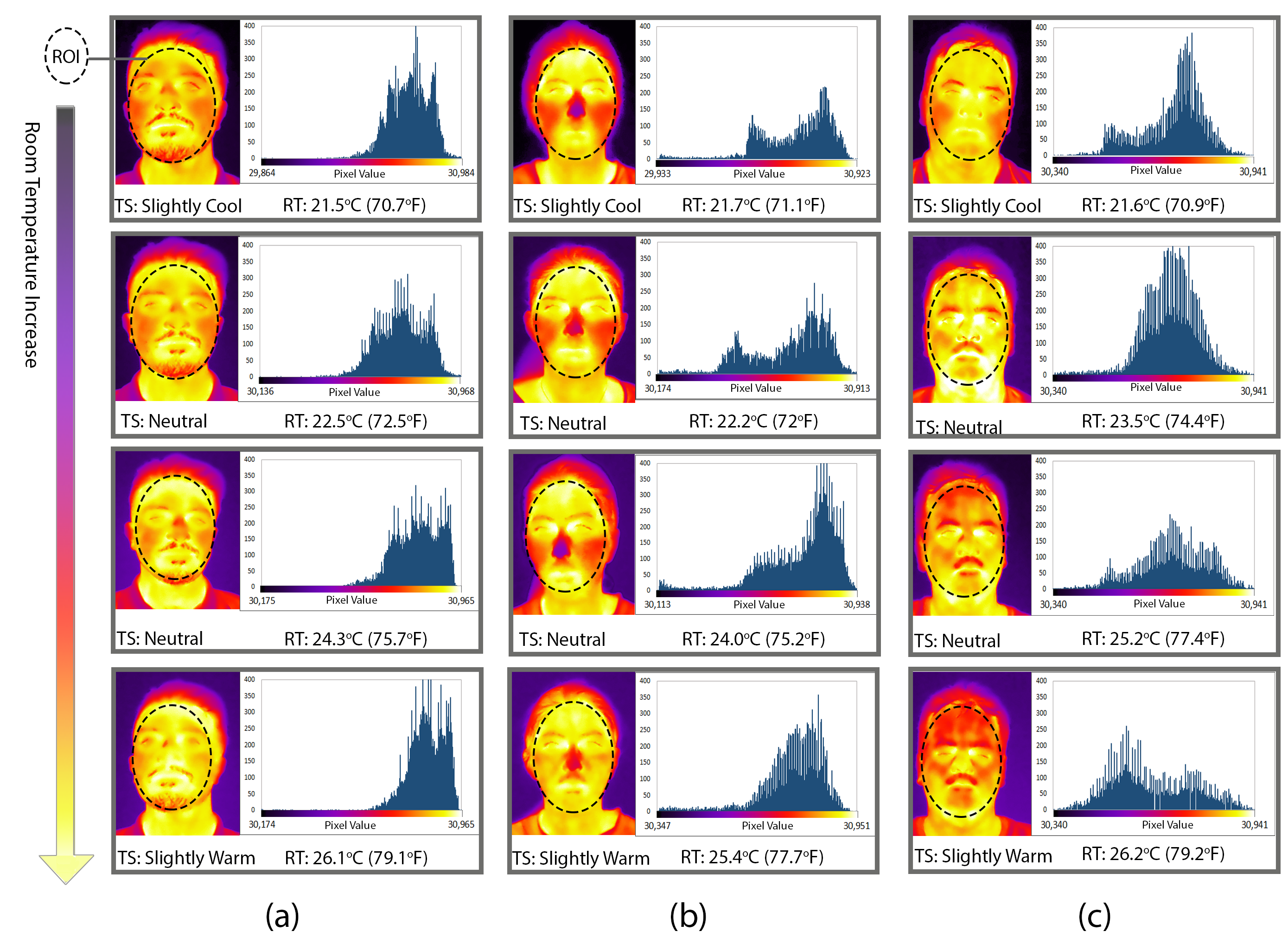}
\vspace{-20pt}
\caption{False Color and Histogram Representation of Sample Data In Different Temperatures, TS: Thermal Sensation, RT: Room Temperature}
\label{fig:histogram1}
\end{figure}

Moreover, as one of the main focuses of this study, the importance of environmental temperature in changing the facial thermal images is highlighted by studying some sample images from the dataset of three selected subjects. Figure \ref{fig:histogram1} shows how the facial skin temperature changes in three different thermal conditions for these three individuals. The first row has the lowest Room Temperatures (RT), which increases gradually as we reach the last row. The RT for each row is approximately the same for all three subjects. In addition, the reported Thermal Sensation (TS) of each subject is included, which is also the same for images of subjects in the same row. Ironbow A false-color pallet is used for easier identification of changes in different facial areas.

The authors have selected an ellipse-shaped Region of Interest (ROI) in the facial area. The histogram of the pixel value distribution in the defined ROI also shows how skin temperature changes on the whole face for each subject. As mentioned in Equation \ref{eq1} the skin temperature is calculated linearly from the pixel value, so the larger amounts of this value show a higher skin temperature. We can observe that the nose and cheek area show more apparent changes than the forehead and chin for all three subjects. In addition, we can see in both the false-colored images and histograms subjects (a) and (b) have a noticeable increase in their skin temperature as the room temperature increases.

The individual differences in facial regions' temperatures can also be detected easily in images of different subjects in the same row when the Room Temperatures (RT) are relatively close together. For instance, in the first row, the room temperature is in the range of (21.5°C-21.7°C), but the user's skin temperature shows apparent differences in different facial regions. It is also important that all three subjects have reported their thermal sensation as "Slightly Cool," while their facial thermal image and the ROI histogram vary to a great extent. While subjects (a) and (b) show an increase in the facial skin temperature, subject (c) shows a different pattern in skin temperature changes, which is an initial increase followed by an identifiable decrease when the temperature reaches 25.2°C (~77°F). The explanation for this behavior change is that the subject was beginning to sweat in the forehead area, which was not noticeable even by himself. The higher ambient temperature was the cause of the sweating. As mentioned by the individual, he typically has a lower tolerance for increased ambient temperatures, which results in him sweating more frequently in a hot room temperature. However, the perspiration was insignificant and therefore not noticeable by the researchers or the subject at the time of the experiment. This behavior highlights the importance of including ambient temperature variation in the facial thermography dataset. \color{black} We can observe that the forehead area of this subject shows an increase when the room temperature changes from 21.6 °C to 23.5 °C and the subject's thermal sensation changes from "Slightly Cool" to "Neutral". However, as the unnoticeable sweating begins in the next two rows, the forehead temperature decreases, while the subject's thermal sensation changes to "Slightly Warm". These individual differences in the facial thermal images have reoccurred in the three next rows, as well as the whole dataset. The personal differences in facial thermal images are one of the main reasons we need a comprehensive dataset in several different thermal conditions for our investigations into facial thermography.

\begin{figure}[t]
\centering
\includegraphics[width=\linewidth]{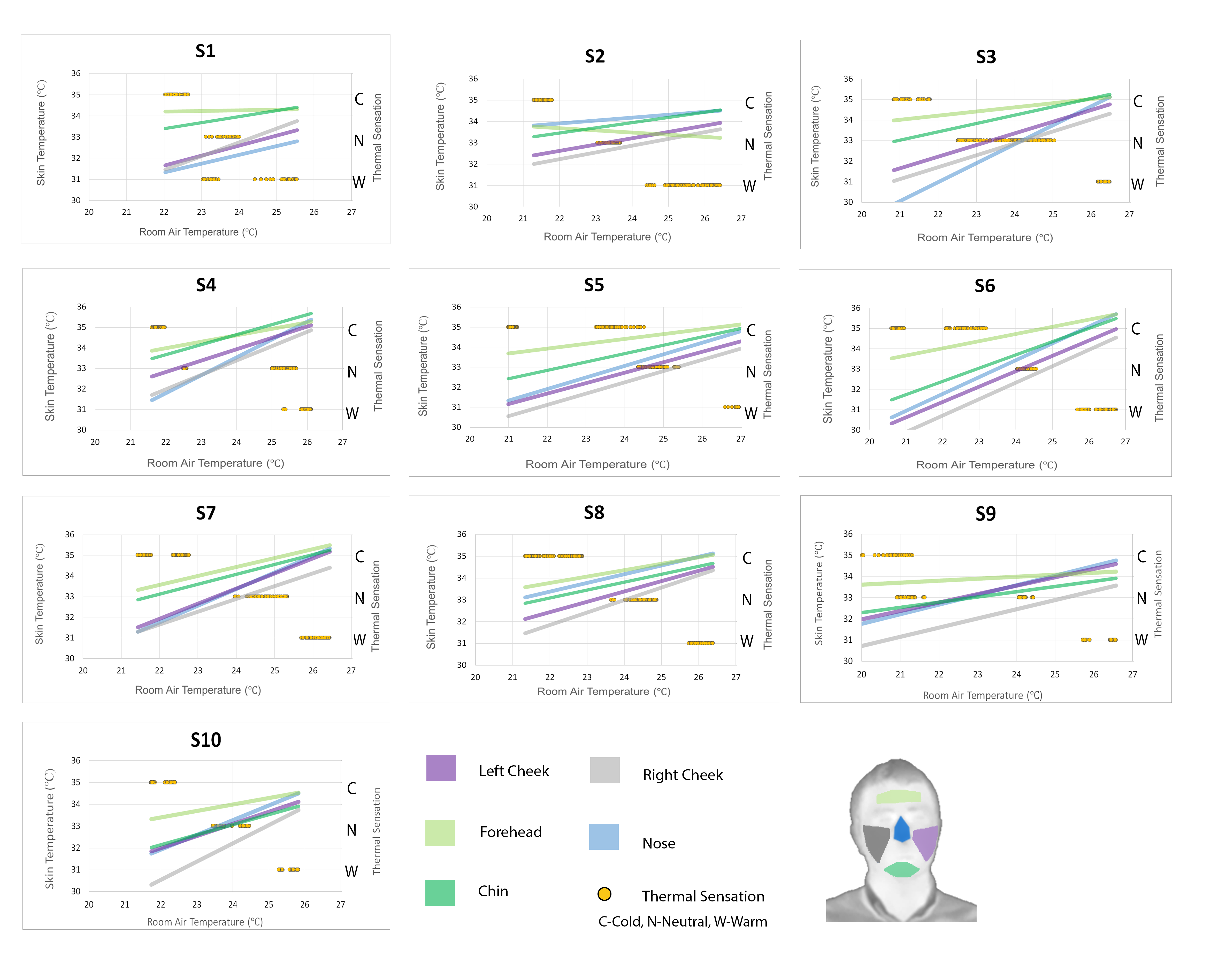}
\vspace{-20pt}
\caption{Linear Correlation of Facial Skin Temperature in Different Areas With Room Temperature For Each Subject In Addition To Their Thermal Sensation (C-Cold, N-Neutral, W-Warm) }
\label{fig:subjects}
\end{figure}

Although the subjects show personal differences in their facial thermal images, we can observe a linear correlation between the temperature of different facial areas and the environmental temperature in all of them. Figure \ref{fig:subjects} presents this linear correlation in all the facial regions and for all the included subjects. The horizontal axis presents the room temperature, and the primary vertical axis shows the facial skin temperature. In addition, each person's reported subjective thermal sensation as the temperature increases is displayed in the secondary vertical axis. As shown in Figure\ref{fig:subjects}, for all the individual subjects, the highest skin temperature variations as the room temperature increases are in the nose and cheeks area, and the lowest variation is in the forehead and chin area. This is important to note that subject 2's forehead temperature decreases as the room temperature increases, which was previously explained in Figure\ref{fig:histogram1} for the same subject in column (c). Another important observation from this figure is the higher temperature differences in facial areas when the room temperature is low compared to higher room temperatures. This pattern is repeated in all the subjects, while the differences are much greater in some of them, such as subjects 3 and 10.  These figures also show that all the subjects have experienced different thermal sensations; however, the pattern and frequency of each thermal vote are different and based on individual differences.

\section{Discussion and Future Work}
This paper presents a  novel facial infrared thermal dataset with variation in environmental properties, distance from the camera, and head position in raw 16-bit data frames. The data is annotated with each person's environmental conditions, facial landmarks, and at the time of recording each frame. The subjective thermal sensation annotations are a new addition to the face thermal image datasets. The comparative study of the temperature in different facial areas shows the importance of environmental temperature in facial thermography. Variation in environmental temperature, facial resolution, and head position makes our dataset an excellent fit for training facial recognition algorithms. Another important consideration is identifying facial landmarks for a more precise evaluation of facial skin temperature. Although we have used the RGB paired images for facial area identification in this paper, there is other state-of-the-art research for facial landmark detection. Since our dataset includes images with various resolutions, it would be a great fit to evaluate the prediction accuracy of such algorithms, which would be one of our future research objectives. The explanation for this behavior change is that the subject was beginning to sweat in the forehead area, which was not noticeable even by himself.

 In the data analysis section for this paper, we have focused on the reliability of the dataset for future research. Therefore, we have looked into the temperature range of the images and the correlation of ambient temperature with the skin temperature of each subject. However, other analyses can be performed on this dataset, including investigation of the performance of different learning algorithms on facial landmark detection. In addition, studies can be performed to investigate the accuracy of detecting subjects' thermal comfort based on their skin temperature. Although previously included research has already proved this possibility \cite{Li2019RobustCameras}, it is important to study this feature in a farther distance from the camera if we want to apply it in real buildings. Another subsequent study would be on the prediction of subjects' thermal sensations based on the infrared thermal images by utilizing the current dataset. We will analyse the influence of distance and head position on the thermal camera readings and the prediction accuracy of thermal comfort based on the subjects' skin temperature. The manual annotation of the facial frames will provide us with the ground truth facial landmarks in the thermal frames, which will be used for the validation of our presented approach. 

More projects are in progress for the improvement of this dataset based on its current limitations. Our dataset does not include different facial expressions, which are included in some datasets and will be added to our updated versions. Currently, due to the high variety of recording conditions for each subject, we have included ten subjects in our experiment. We are planning on adding more subjects in the future version of our dataset from different ethnic groups. The current dataset was recorded in a controlled indoor environment. Including images that were taken outdoors could make this dataset more suitable applicable. In addition, our current dataset does not include RGB images, which can be included in our following versions to make the dataset helpful for a broader range of projects.

The complete dataset is available for downloading on our Github page \cite{TeCSAR-UNCC/UNCC-ThermalFace}.
\\
\color{black}
\textbf{Funding Sources}\\
This work was supported by the National Science Foundation [grant number NSF 2104223].
\bibliography{3.bib,references1.bib}
\end{document}